\documentclass{fairmeta}
\usepackage{amsmath}
\usepackage{enumerate} 
\usepackage{bm}
\usepackage{algorithm}
\usepackage{floatflt}
\usepackage{algpseudocode}
\usepackage{amsfonts}
\usepackage{amsthm}
\usepackage{newtxtt}
\usepackage{algorithm}
\usepackage{colortbl} 
\usepackage{cleveref}
\usepackage{diagbox} 
\usepackage[utf8]{inputenc}
\usepackage{textgreek}
\usepackage{colortbl}
\usepackage{nicematrix}
\usepackage{makecell}
\usepackage{float}
\usepackage{arydshln}
\usepackage[frozencache,cachedir=.]{minted}
\usepackage{caption}
\usepackage{subcaption}
\usepackage{tcolorbox}
\usepackage{amssymb}
\usepackage{xspace}
\usepackage{wrapfig}
\usepackage{adjustbox}
\usepackage{tabularx}
\usepackage{booktabs}
\usepackage{mathtools}
\usepackage{wrapfig}
\usepackage{amssymb}
\usepackage{graphicx}

\usepackage{silence}
\makeatletter
\patchcmd{\wrong@fontshape}{\@gobbletwo}{}{}{}
\makeatother
\definecolor{upColor}{RGB}{17,138,21}
\definecolor{downColor}{RGB}{174,36,67}

\newtheorem{theorem}{Theorem}[]

\newtheorem{remark1}[theorem]{Remark}

\title{Ruyi2 Technical Report}
\author[]{Huan Song}
\author[]{Shuyu Tian}
\author[]{Junyi Hao}
\author[]{Minxiu Xu}
\author[]{Hongjun An}
\author[]{Yiliang Song}
\author[]{Jiawei Shao}
\author[]{Xuelong Li}
\affiliation[]{Institute of Artificial Intelligence (TeleAI), China Telecom}

\date{
   February 25, 2026 
   
\textbf{Model:} \url{https://huggingface.co/TeleAI-AI-Flow/AI-Flow-Ruyi2} \\
    \textbf{ Code:} \ \ \url{https://github.com/TeleAI-AI-Flow/AI-Flow-Ruyi2}
}

\metadata[Correspondence to]{Xuelong Li (\email{xuelong\_li@ieee.org})}

\begin{document}

\abstract{
Large Language Models (LLMs) face significant challenges regarding deployment costs and latency, necessitating adaptive computing strategies. Building upon the AI Flow framework, we introduce Ruyi2 as an evolution of our adaptive model series designed for efficient variable-depth computation. While early-exit architectures offer a viable efficiency-performance balance, the Ruyi model and existing methods often struggle with optimization complexity and compatibility with large-scale distributed training. To bridge this gap, Ruyi2 introduces a stable "Familial Model" based on Megatron-LM. By using 3D parallel training, it achieves a 2–3× speedup over Ruyi, while performing comparably to same-sized Qwen3 models. These results confirm that family-based parameter sharing is a highly effective strategy, establishing a new ''Train Once, Deploy Many'' paradigm and providing a key reference for balancing architectural efficiency with high-performance capabilities.
}

\maketitle

\section{Introduction}

Large Language Models (LLMs) have demonstrated remarkable capabilities across a wide range of natural language understanding and generation tasks \citep{brown2020language,bommasani2021opportunities}. However, their rapidly growing parameter scales and inference costs pose significant challenges for practical deployment, particularly in scenarios constrained by latency, throughput, and energy consumption. While Scaling Laws \citep{kaplan2020scaling,hoffmann2022training} indicate that performance improves predictably with increases in model size and data, real-world applications often necessitate ``adaptive computing,'' where simple inputs do not require invoking the full capacity of the largest model \citep{schuster2022confident,jiang2024d,manvi2024adaptive}.

Early-exit mechanisms \citep{teerapittayanon2016branchynet} have emerged as an effective solution to this challenge \citep{chen2023ee,elhoushi2024layerskip,valade2024accelerating}. By incorporating auxiliary output heads within the intermediate layers of transformer models \citep{xin2020deebert,liu2020fastbert}, LLMs can terminate inference prematurely upon reaching sufficient confidence, thereby reducing computational costs without the need to retrain multiple independent models. This paradigm effectively transforms a single backbone into a family of nested sub-models with varying depths and computational costs \citep{elbayad2019depth}. In this paper, we term such architectures Familial Models and treat them as equivalent to early-exit Large Language Models \citep{schuster2022confident}.

Despite its conceptual simplicity, the large-scale training and deployment of Familial Models face multiple challenges. Naive multi-exit training can significantly exacerbate optimization complexity due to gradient interference between different exits across layers. Furthermore, large-scale LLM training typically relies on pipeline, tensor, and data parallelism, and existing early-exit methodologies often exhibit incompatibility with these distributed training mechanisms. Moreover, incorporating instruction tuning into multi-exit settings introduces new questions regarding supervision placement, loss balancing, and parameter efficiency. While approaches such as HELIOS \citep{kumar2025helios} and Balcony \citep{jamialahmadi2025balcony} have explored architectural improvements to enhance inference adaptability, and Dynamic Speculative Decoding \citep{chen2023accelerating} has been investigated for efficiency, gaps remain. Regarding model scaling, \cite{kim2024solar}\ proposed a simple yet effective depth up-scaling technique (Solar 10.7B) \citep{kim2024solar} to expand LLMs, while LLaMA-Pro \citep{wu2024llama} utilized Block Expansion to efficiently increase capacity. However, most existing methods either focus on small-to-medium scale models or overlook the system-level challenges associated with efficiently training billion-parameter Familial Models. Additionally, while standard Familial Models and SVD-LLM v2 \citep{wang2025svd} have proposed decomposition methods, integrating these into a coherent post-training pipeline remains complex.

Building upon the AI Flow theoretical framework \citep{AIFlowattheNetworkEdge,Task-Oriented}, we introduce Ruyi2 for efficient variable-depth computation. Although early-exit architectures offer a viable efficiency-performance trade-off, prior implementations such as Ruyi \citep{an2025aiflowperspectivesscenarios} exhibited significant limitations regarding optimization complexity and integration with large-scale distributed systems. To resolve these issues, Ruyi2 implements several critical advancements. Primarily, it introduces a stable Familial Model architecture based on Megatron-LM. By integrating 3D parallel training\citep{narayanan2021efficient}, the framework is systematically optimized, achieving a 2–3x speedup in training efficiency. Consequently, a single training session yields multiple deployable sub-models, realizing a ``Train Once, Deploy Many'' paradigm \citep{cai2019once,hou2020dynabert}. Furthermore, through the rigorous refinement of training data and the application of multi-stage training strategies, the Ruyi2 series consistently outperforms its predecessor, Ruyi, while demonstrating capabilities comparable to same-sized Qwen3 \citep{yang2025qwen3} models.

This technical report systematically outlines the training of Familial Models, emphasizing instruction tuning, model architecture, and the training framework. We first discuss how to integrate instruction supervision into multi-exit models under both full-parameter and parameter-efficient regimes. Subsequently, we describe the architectural design space of Familial Models, including exit placement, parameter sharing, and head specialization. Finally, we elaborate on the scalable training framework suitable for Familial Models, highlighting how modern distributed systems enable large-scale efficient optimization. To further enhance the capabilities of specific edge-side nodes, we propose the DaE (Expand-then-Decompose) framework for the $1.7\text{B}$ variant. Unlike layer cloning strategies commonly used in previous depth scaling work, we employ Stable Block Expansion (SBE) with random internal initialization to avoid optimization inertia and representation collapse. We then utilize a post-training compression stage derived from SVD-LLM v2 techniques, reducing the parameter count by $40\%$ with only marginal performance degradation.

Experimental results demonstrate that the proposed method achieves excellent performance on reasoning-intensive benchmarks such as GSM8K and MATH. This further verifies the adherence of the familial architecture to scaling laws, proving that the sharing mechanism does not disrupt the backbone's scaling behavior. This achievement is attributed to the 3D-parallel compatible Familial Models training framework, which resolves the gradient interference challenge of large-scale joint optimization via customized loss aggregation and pipeline-aware backpropagation, enabling full-parameter ``zero-overhead'' training. For edge-side deployment, this work introduces the DaE paradigm, utilizing zero-residual random initialization to overcome optimization inertia during model expansion, combined with SVD compression to reduce incremental parameters by $40\%$ while keeping performance loss within $2\%$. Augmented by a multi-stage alignment strategy involving ``Continual Pre-training -- Fine-tuning -- Reinforcement Learning'' (incorporating GRPO), this scheme significantly enhances the reasoning capabilities of small-scale sub-models on complex logic tasks, bridging the gap with the backbone model. Ultimately, the system realizes an efficient ``Train Once, Produce Many'' paradigm, providing a valuable technical pathway for achieving adaptive high-performance inference in resource-constrained environments.

\section{Architecture}

Consistent with the Ruyi architecture\citep{an2025aiflowperspectivesscenarios}, Ruyi2 Familial Models use a \emph{shared backbone} and multiple \emph{exit heads}. 
The backbone is a standard transformer block. 
The family-model aspect arises from attaching lightweight output modules at selected depths. 
For example, one may select layers 6 and 12 to host exit heads.
Each exit head $\phi_i$ takes the hidden state $\mathbf{x}_i$ from layer $i$ and projects it to a token distribution $\mathbf{o}_i$. 
Ruyi2 Familial Models is designed as follows: Ruyi2-1.7B is a sub-model using layers 1-3, Ruyi2-8B is a sub-model using layers 1-22, and Ruyi2-14B uses the full model with all 40 layers.
Notably, all sub-models share the parameters of the common layers, so no separate models need to be trained from scratch. The architecture incorporates a nested early-exit design, where Decomposed Blocks are distributed across various depths of a shared Transformer backbone. This design enables the model to flexibly complete inference and generate Branch Outputs across End devices, Edge servers, or Cloud servers, depending on the available computational resources and task difficulty. 

Ruyi2 is built upon Qwen3-14B-Base \citep{yang2025qwen3}, with its largest branch having 14B parameters and supporting early-exit branches with effective parameter sizes of 1.7B and 8B. The 14B main branch was initialized directly from the base model parameters; for the 8B and 1.7B early-exit branches, their decoder layers were initialized using parameters from the layer immediately following their respective early-exit positions. After initialization, we employed a multi-branch joint pre-training approach, conducting continued pre-training on approximately 800 billion tokens from a proprietary high-quality dataset to construct Ruyi2-Base. Subsequently, we performed joint instruction-following fine-tuning on all branches using approximately 4 million high-quality instruction examples, resulting in Ruyi2. Ruyi2 comprises three parameter-sharing models: Ruyi2-14B, Ruyi2-8B, and Ruyi2-1.7B.
\begin{table}[h]
    \centering
    
    \begin{tabular}{lccccc}
        \toprule
        \textbf{Model} & \textbf{Layers}   &\textbf{Hidden Size}& \textbf{Tie Embedding} & \textbf{Heads (Q / KV)} & \textbf{Context Length} \\
        \midrule
        Ruyi2--1.7B & 3  &5120& No & 40 / 8 & 8K \\
 Ruyi2--8B& 22 &5120& No& 40 / 8&8K\\
 Ruyi2--14B& 40 &5120& No& 40 / 8&8K\\
 \bottomrule
    \end{tabular}
\captionsetup{justification=centering} 
    \caption{Model architecture details for Ruyi2.}
\end{table}

\section{Pre-training}
\subsection{Data Acquisition}
To build a high-performance foundation model with strong reasoning, coding, and general language capabilities, we constructed a massive, carefully curated pre-training corpus that emphasizes high-density quality over raw web scale. The dataset spans multilingual web text, high-fidelity scientific literature, executable code, and symbolic mathematics. Noisy web data are rigorously filtered and globally deduplicated to retain only information-rich content. Unstructured knowledge from PDFs (e.g., papers, textbooks, manuals) is extracted via a vision-centric pipeline based on fine-tuned vision–language models, followed by a refinement loop using a powerful language model to correct OCR artifacts and restore semantic coherence. Code and math data are sourced from well-structured open-source projects and authoritative academic materials, preserved in original formats (e.g., LaTeX) to maintain structural integrity. Finally, high-quality synthetic data are introduced during mid-training to enhance complex reasoning and general problem-solving abilities.

\subsection{Data Processing}
\begin{figure}[h]
    \centering
    \includegraphics[width=1\linewidth]{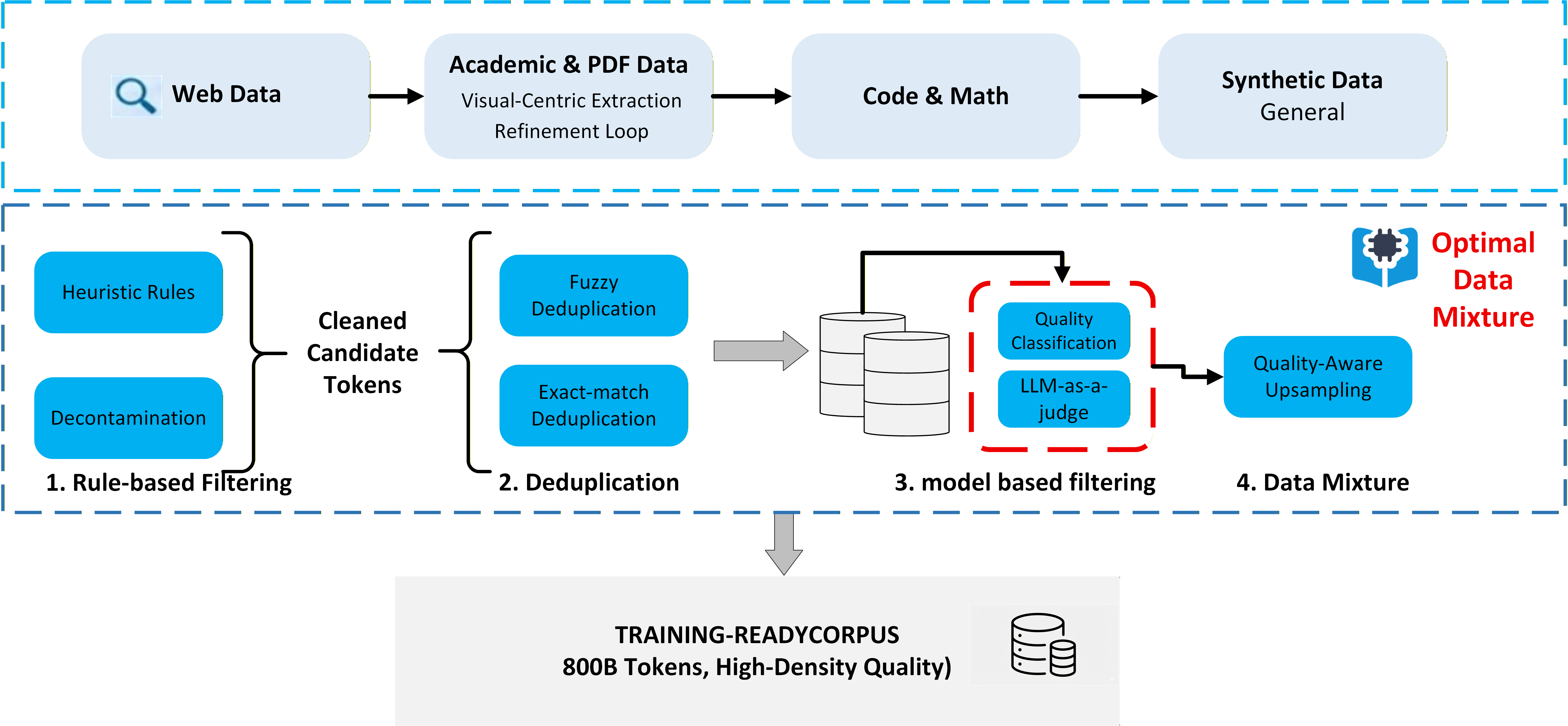}
    \caption{\textbf{The data engineering pipeline for constructing a high-quality training-ready corpus.} The pipeline consists of four main stages: filtering, deduplication, decontamination, and data mixture, ensuring the high density and quality of the dataset.}
    \label{fig:placeholder666}
\end{figure}

Transforming raw data into a training-ready corpus relies on an industrial-grade, multi-stage processing pipeline that prioritizes token quality while preserving semantic diversity. As illustrated in Figure \ref{fig:placeholder666}, the pipeline integrates multi-level filtering—combining fast heuristic rules with model-based classifiers—to remove low-quality, repetitive, non-target-language, and spam content. To prevent memorization and improve efficiency, we apply advanced multi-granular deduplication, including MinHash/LSH-based fuzzy matching and high-precision suffix-array–based substring deduplication, which is especially effective for code and academic text. Finally, the data mixing strategy is rigorously optimized to establish source weights that balance diversity and complexity. This is coupled with quality-aware upsampling to deliberately increase the exposure of verified, high-value data, enabling higher information density without compromising generalization.
\subsection{Pre-training Strategy}
\subsubsection{Multi-stage Continual Pre-Training Strategy}
Our pre-training pipeline employs a dynamic, multi-stage curriculum learning strategy that progressively adjusts data composition and optimizer hyperparameters, allowing the model to evolve smoothly from general language modeling to high-level task proficiency. The process begins with General Pre-training, where the model establishes a robust representation space and broad world knowledge by ingesting a large-scale corpus of approximately [NUM] tokens. This provides a solid semantic foundation for the subsequent Mid-training \& Annealing phase, which shifts the paradigm toward domain specialization by significantly up-sampling high-quality STEM, code, and high-fidelity synthetic reasoning data. To accommodate this drastic shift in data distribution, we apply a learning rate annealing strategy toward the end of this phase; this smooth decay trajectory mitigates gradient instability and prevents catastrophic forgetting of general representations, ensuring stable convergence and a robust initialization for downstream fine-tuning. 
\subsubsection{Joint Optimization Framework}
We introduce a joint training paradigm in which the main backbone model and all familial branches are optimized simultaneously. This approach ensures that the shared intermediate layers acquire representations that remain robust for both the full-depth model and the early-exiting branches. Formally, given a batch of input tokens, the total training objective $\mathcal{L}_{\text{total}}$ is defined as a weighted sum of the cross-entropy losses from the main backbone and the $K$ auxiliary branches:

\begin{equation}
    \mathcal{L}_{\text{total}} = \sum_{k=1}^{K} \lambda_k \mathcal{L}_{\text{branch}\_k}
\end{equation}

where $\mathcal{L}_{\text{branch}\_k}$ represent the causal language modeling loss for the final layer and the $k$-th intermediate exit, respectively. The term $\lambda_k$ denotes a time-dependent weighting coefficient that modulates the influence of auxiliary branches throughout the training process. This scheduling is crucial for providing deep supervision to lower layers in the early stages to accelerate convergence, while progressively decaying effectively to prevent the auxiliary objectives from conflicting with the global optimization of the main backbone during the final convergence phase.
\section{Post-training}
\subsection{Overall Post-training Pipeline}
\begin{figure}[h]
    \centering
    \includegraphics[width=1\linewidth]{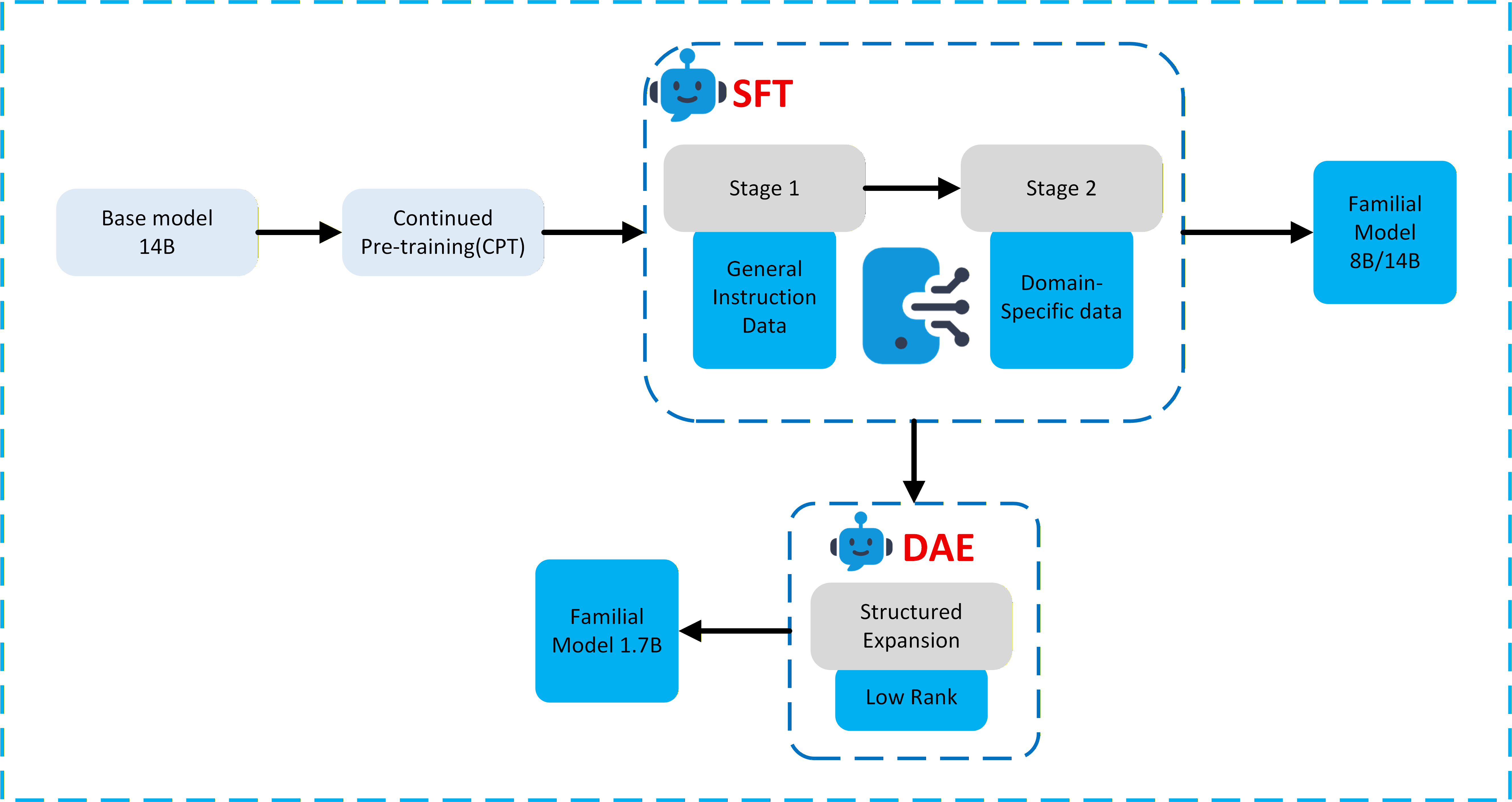}
    \caption{Overview of the Staged Capability Improvement Pipeline}
    \label{fig:placeholder667}
\end{figure}

As illustrated in Figure \ref{fig:placeholder667}, the overall training pipeline introduced in this paper aims to construct a comprehensive family of models, supporting staged capability improvement and coordinated evolution across models of different scales. 
First, we utilize large-scale domain-specific corpora for \textit{Continued Pre-training (CPT)} to facilitate the transfer from general knowledge to domain knowledge and unify the representation spaces across models of different scales. 
Second, we adopt a \textit{multi-stage instruction alignment strategy}, which first strengthens fundamental capabilities via general instruction fine-tuning and subsequently introduces domain data to enhance professional performance, ensuring a balance between versatility and specialization. 
Finally, for small-scale models, we implement an ``expand-then-compress'' incremental parameter learning strategy: incremental parameters are introduced during training to boost representation capacity and then reduced by $40\%$ using low-rank decomposition before inference. This approach preserves high performance without increasing inference overhead, thereby enabling efficient edge deployment.

\subsection{Post-training Data}
\begin{figure}[h]
    \centering
    \includegraphics[width=1\linewidth]{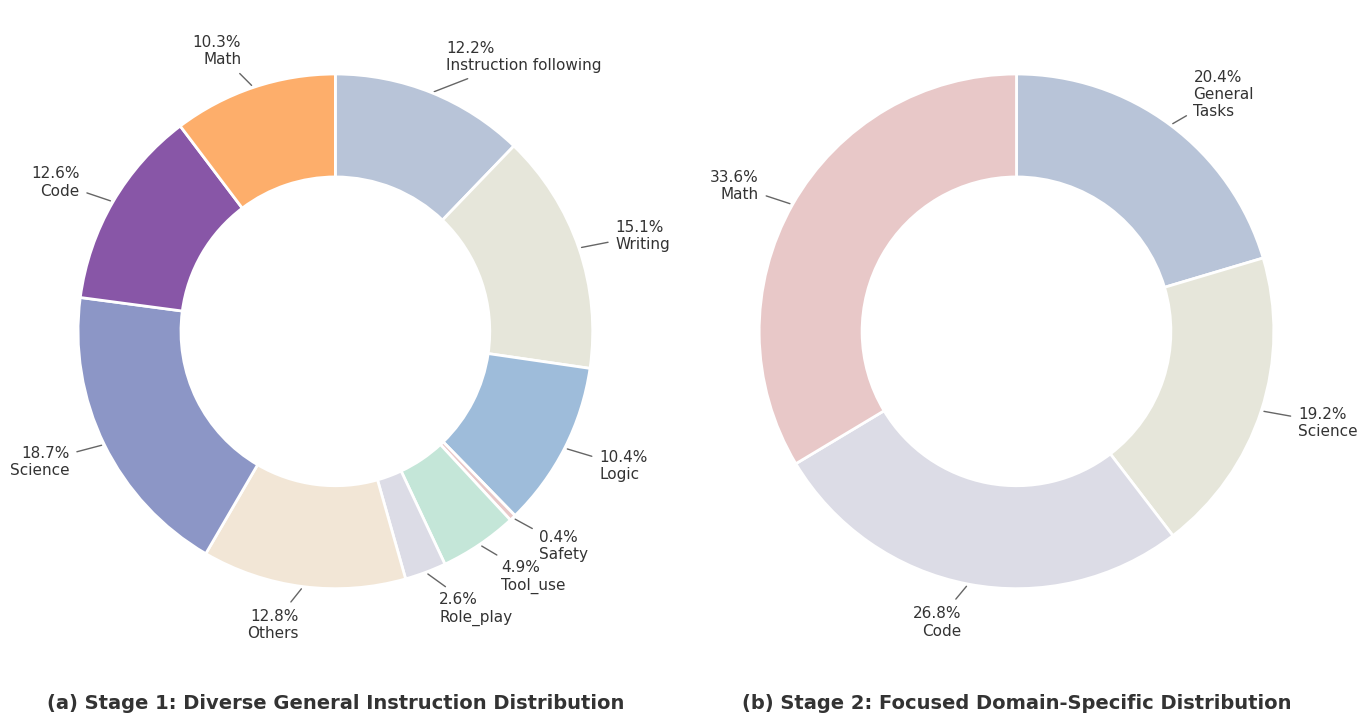}
    \caption{SFT Data Mixture Strategy: Evolution from Diverse Generalization to Focused Domain Specialization }
    \label{fig:placeholder1112}
\end{figure}

Post-training Data During the SFT stage, this study abandons traditional coarse-grained fine-tuning and instead deploys a sophisticated data-engineering system aimed at constructing a high-quality, high-density instruction dataset. The system covers the entire pipeline, including hierarchical categorization, instruction evolution, and multi-dimensional data cleaning. For data curation, we establish a three-level instruction labeling taxonomy spanning over 20 vertical domains. Leveraging high-capability teacher models, we implement Synthetic Instruction Evolution, in which simple instructions are upgraded into high-difficulty composite samples by explicitly introducing complex constraints. Meanwhile, to ensure data purity and safety, we apply a series of stringent filtering and validation procedures, including heuristic rule–based filtering, rigorous decontamination against downstream benchmarks (e.g., GSM8K and HumanEval), and a model-based quality scoring mechamism. 
For data-mixture decisions, we move away from blind empirical guesswork and adopt a scientific ratio-selection strategy based on proxy models. By conducting extensive ablation studies on small-scale 3B and 7B models, we fit spline curves that characterize the impact of different data categories. This enables the principled determination of optimal mixing ratios. This approach moves from empirical tuning to quantitative optimization. It yields scalable data-mixture strategies that significantly enhance capabilities in mathematics, coding, reasoning, and dialogue. As shown in Figure \ref{fig:placeholder1112}, SFT data construction is divided into two stages: Stage 1 focuses on diversity (Instruction following, Others), while Stage 2 significantly increases the proportion of Math and Code to enhance specific capabilities. 

\subsection{Post-training Strategy}

\subsubsection{Multi-stage Instruction Alignment}
To address the ``capability elicitation'' bottlenecks inherent in standard supervised fine-tuning (SFT), this study proposes a phased alignment strategy based on curriculum learning principles. Distinct from the knowledge accumulation focus of the pre-training phase, this strategy progressively guides the model from establishing foundational capabilities to mastering specialized domain expertise.

\begin{enumerate}
    \item \textbf{Stage I: General Behavioral Alignment.}
    This stage aims to build instruction-following capabilities, ensuring consistency and compliance in model interactions. By leveraging diverse data covering multi-turn dialogues and wide-ranging general instructions, the training prioritizes enhancing intent alignment, response precision, and output formatting standards. Simultaneously, it establishes safety boundaries to lay a solid general base for the subsequent expansion of specific task capabilities. 

    \item \textbf{Stage II: Domain Capability Injection.}
Building upon general alignment, this stage shifts the optimization focus toward specialized knowledge depth in vertical domains and complex task-solving capabilities. The training data is rigorously curated, encompassing STEM, programming, and tasks with specific constraints. To mitigate the risk of catastrophic forgetting regarding general capabilities during domain adaptation, an Experience Replay mechanism is employed. High-quality general instructions from Stage I are integrated into the training process at a fixed ratio, ensuring the preservation of generalizability while enhancing domain expertise.
\end{enumerate}

\subsubsection{1.7B Architecture Specific Optimization}
Within our Ruyi2 Familial Models, the 1.7B variant serves as a key node for edge-side deployment, offering a balance between inference latency and generation quality. Nevertheless, due to its reduced depth, this branch has ability constraints in complex semantic comprehension and performing in reasoning tasks when compared to larger familial variants. Addressing these limitations in an efficient and reasonable manner, without training from scratch, remains an important challenge.

\begin{figure}[h]
    \centering
    \includegraphics[width=0.75\linewidth]{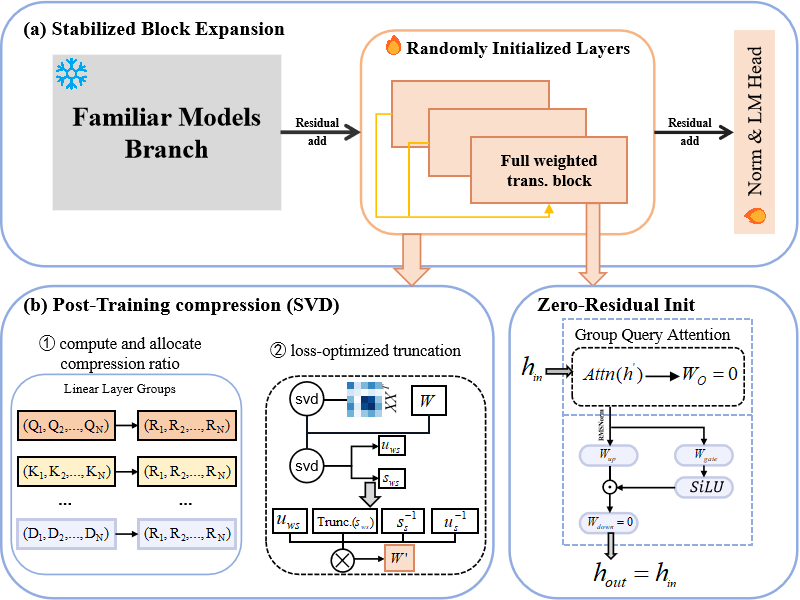}
    \caption{Two stages of DaE}
    \label{fig:placeholder78}
\end{figure}

\subparagraph{Architecture Innovation: The DaE Framework}
To this end, we propose a two-stage framework termed DaE (Decompose after Expansion), as illustrated in Figure \ref{fig:placeholder78}, which integrates structured model expansion with post-training compression.
In the first stage, inspired by recent advances in efficient scaling methods such as LLaMA-Pro and Solar, we apply a stabilized block expansion strategy to an early-exit branch of the familial model. Specifically, several full-parameter Transformer blocks are appended to the branch, while the original model parameters are frozen and only the newly added blocks are optimized. This design allows the expanded model to fully exploit its representational capacity, thereby achieving enhanced performance prior to subsequent compression.

In the second stage, to control the parameter budget and improve deployment efficiency, we apply low-rank decomposition to the newly introduced blocks. Specifically, singular value decomposition (SVD) is employed to re-parameterize the trained full-parameter Transformer blocks into compact low-rank representations. By carefully tuning the compression ratio, this decomposition process substantially reduces both the parameter count and computational overhead, while preserving the majority of performance gains obtained from the block expansion stage. Overall, the proposed two-stage design achieves an effective trade-off between model capacity enhancement and efficiency constraints.
\subparagraph{Stabilized Block Expansion}

Following the methodology of Kim et al., we upscale the depth of the 1.7B branch while preserving its original feature space. Concretely, we append $N=3$ additional Transformer blocks to the exit of the branch. A central challenge in such block expansion—where newly initialized layers are grafted onto an existing backbone—is maintaining training stability. Naïvely optimizing the expanded model often leads to unstable gradients or representation collapse during the early stages of training.

To address this issue, we adopt a zero-residual initialization strategy. Let $h_L$ denote the hidden state at the exit layer $L$ of the original backbone. For the $k$-th newly added block ($1 \leq k \leq N$), the forward computation is defined as:
\[
h_{L+k} = h_{L+k-1} + \mathrm{Block}_k(h_{L+k-1}),\tag{2}
\]

we impose a strict initialization constraint on the final projection matrices of both the attention and MLP submodules:
\[
W_o = 0, \quad W_{\text{down}} = 0,\tag{3}
\]
where $W_o$ denotes the output projection matrix of the grouped-query attention (GQA) module, and $W_{\text{down}}$ represents the down-projection matrix of the MLP.
As a result, at initialization time ($t=0$),
\[
\mathrm{Block}_k(h) = 0 \quad \Rightarrow \quad h_{L+k} = h_{L+k-1},\tag{4}
\]
which causes the expanded branch to degenerate into an exact identity mapping of the original exit representation. This initialization scheme preserves the original feature distribution while enabling stable gradient propagation through the newly introduced depths during early training, thereby ensuring a smooth and stable optimization process.

\begin{figure}[h]
        \begin{minipage}{0.45\textwidth}
                \centering
                \includegraphics[width=1\textwidth]{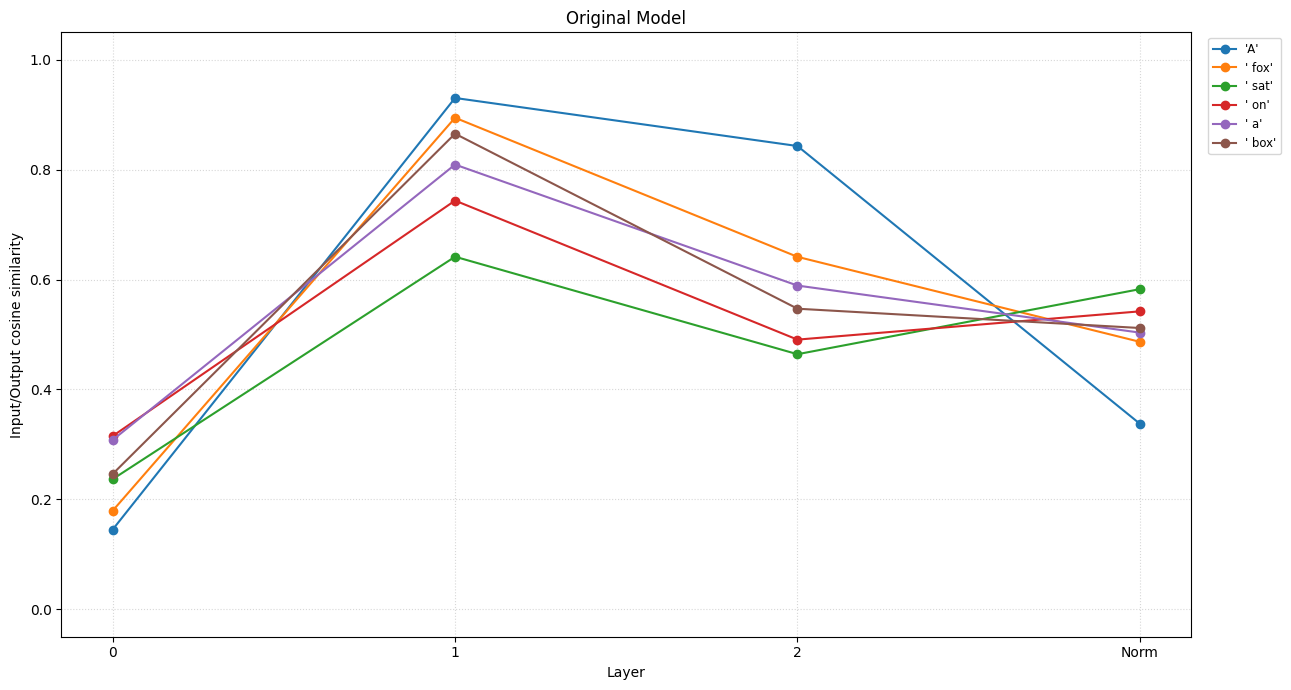}

         \end{minipage}
         \begin{minipage}{0.51\textwidth}
                \centering
                \includegraphics[width=0.9\textwidth]{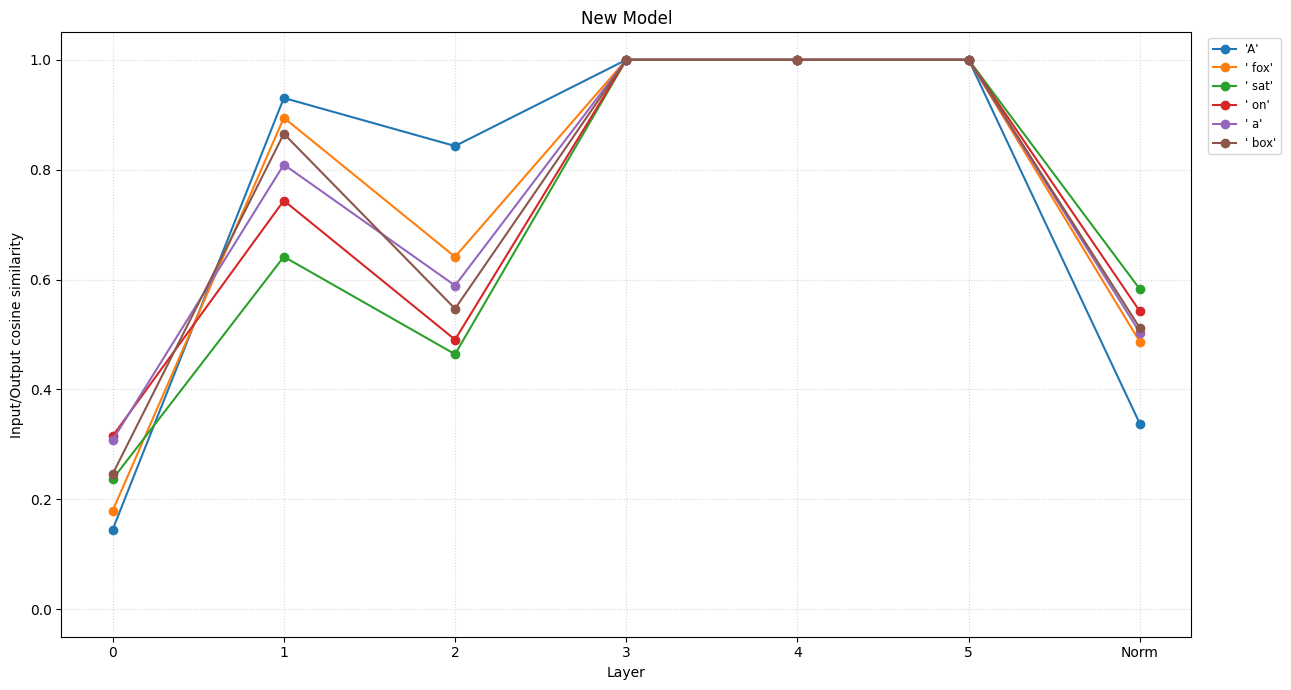}

         \end{minipage}
\caption{Input/output cosine similarity score of each token per layer for text "A fox sat on a box". }
\label{QA}
\end{figure}
A second critical design choice concerns the initialization of the internal parameters of the newly added blocks. Prior works on depth expansion, such as LLaMA-Pro, commonly adopt layer cloning, where weights from adjacent pretrained layers are duplicated to initialize new blocks. While effective at larger scales, our ablation studies at the 1.7B scale reveal that this strategy introduces pronounced \textit{optimization inertia}: the induced parameter symmetry causes the new blocks to largely replicate existing feature extraction patterns, thereby limiting their capacity to learn novel representations.

To overcome this limitation, we adopt \textit{randomized internal initialization}. Specifically, the internal weights of each new block, namely $(W_Q, W_K, W_V, W_{\text{gate}}, W_{\text{up}})$, are initialized using Gaussian random distributions rather than being cloned from the backbone. When combined with the zero-output constraint described above, this design maximizes the structural entropy of newly added layers while maintaining global stability. Intuitively, random internal initialization prevents representation redundancy, while zero-initialized residual paths reduce the impact of noise on the backbone network.
\begin{figure}[h]
    \centering
    \includegraphics[width=1\linewidth]{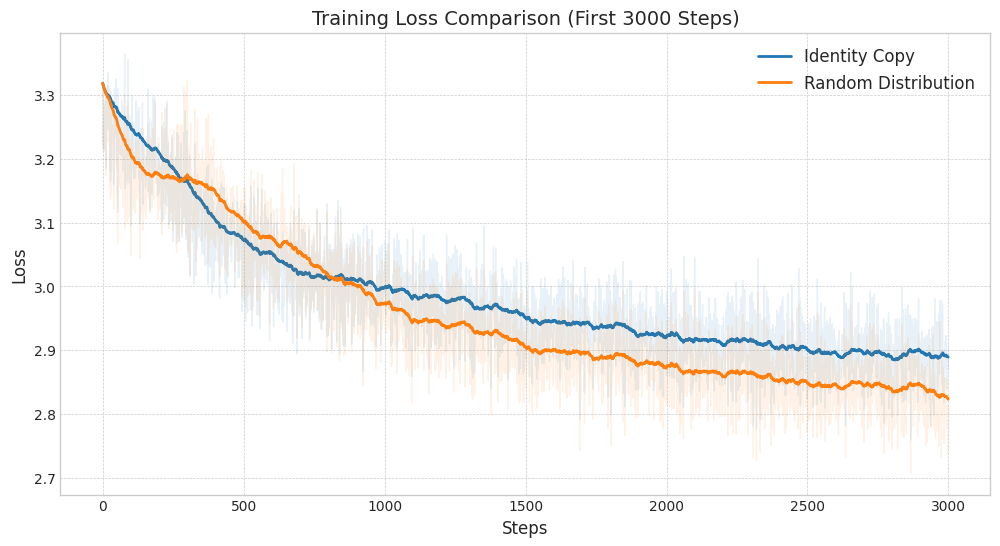}
    \caption{Training loss dynamics on the new model (0-3k steps). The Layer Cloning baseline (red) exhibits optimization inertia with a slower descent rate. In contrast, our Randomized Internal Initialization (purple) consistently achieves lower loss values , demonstrating superior convergence efficiency and a stronger capacity for learning novel representations without being constrained by the backbone's existing patterns.}
    \label{fig:placeholder}
\end{figure}
These two mechanisms enable the newly added blocks to anchor their optimization to the original 1.7B model as a stable reference point, while simultaneously expanding the model’s functional capacity beyond its original depth. This design effectively decouples capacity expansion from representation perturbation, allowing the model to acquire complementary and higher-order reasoning behaviors without inducing catastrophic interference with existing knowledge structures.

\paragraph{Post-Training Low-Rank Decomposition.}
While the expanded blocks significantly enhance the model’s capacity for complex reasoning, the resulting increase in parameter count introduces substantial challenges for resource-constrained deployment scenarios. In particular, mobile and edge devices typically operate under strict memory and latency constraints, with available system memory often limited to single-digit gigabyte levels after accounting for runtime buffers and auxiliary services. As a result, directly deploying the expanded model in its full-parameter form becomes impractical.
To strictly comply with the memory budgets of mobile edge environments (e.g., a $\sim$4--8~GB RAM budget for model weights and inference buffers), we perform SVD-based weight decomposition on both the newly expanded Transformer blocks and the branch-specific output heads during the second stage of the DaE framework. This compression step is designed to preserve the majority of the performance gains introduced by depth expansion while substantially reducing memory footprint and computational overhead.
Unlike the decomposition described in standard Familial LLMs, our approach is applied as a post-training refinement. 
We adopt the technique proposed in SVD-LLM v2 to minimize the reconstruction error of layer activations.

(1) \textbf{Determining the compression ratio.}  
We first determine the compression ratio for each layer. A small calibration set is used to compute the input activations $\mathbf{X}$. The theoretical truncation loss is defined as the reconstruction error of the layer output:
\begin{equation}
\mathcal{L}_{\min} = \left\lVert \mathbf{W}\mathbf{X} - \mathbf{W}'\mathbf{X} \right\rVert_F ,\tag{5}
\end{equation}
We then apply $1 / \log(\mathcal{L}_{\min})$ for inversion and normalization. Finally, the compression ratio for each weight matrix within a group is determined based on the target model compression ratio $R$ and the grouping of weight matrices:
\begin{equation}
\text{Ratio} = \mathrm{Len}(L_G) \times R \times \frac{\mathcal{L}_{\min}}{\mathrm{Sum}(L_G)},\tag{6}
\end{equation}
where $L_G$ denotes the list of theoretical truncation losses for all matrices within the same group, $\mathrm{Len}(L_G)$ denotes the group size, and $\mathrm{Sum}(L_G)$ denotes the sum of losses within the group.

(2) \textbf{Computing the compressed weight matrix.}  
We then compute the compressed weight matrix $\mathbf{W}'$. First, from the input activation covariance $\mathbf{X}$, we construct the whitening matrix $\mathbf{S}$ by applying Cholesky decomposition on $\mathbf{X}\mathbf{X}^\top$. To ensure numerical stability, we perform SVD on the covariance matrix $\mathbf{X}\mathbf{X}^\top$ to obtain $\mathbf{U}_s$, $\mathbf{S}_s$, and $\mathbf{V}_s$. Subsequently, $\mathbf{U}_{ws}$, $\mathbf{S}_{ws}$, and $\mathbf{V}_{ws}$ are obtained by performing SVD on $\mathbf{W} \times \mathbf{U}_s \times \sqrt{\mathbf{S}_s}$. Finally, the decomposed weight matrix is given by:
\begin{equation}
\mathbf{W}' = \mathbf{U}_{ws} \times \mathrm{Trunc.}(\mathbf{S}_{ws}) \times \mathbf{V}_{ws} \times \sqrt{\mathbf{S}_s^{-1}} \times \mathbf{U}_s^{-1}.\tag{7}
\end{equation}
\subsection{Experiment}
\subsubsection{Experimental Setup}
Leveraging the 1.7B branch of our established LLM as the backbone, we implement the SBE strategy by appending N=3 additional Transformer blocks initialized via zero-residual and randomized internal methods; this architecture effectively doubles the total parameter count while maintaining approximately half as trainable parameters. Regarding datasets, we facilitate domain adaptation by pretraining on a 600B-token corpus dominated by STEM and reasoning-intensive texts, supplemented with 10\% general replay data to mitigate catastrophic forgetting. Subsequently, the model undergoes Supervised Fine-Tuning (SFT) on roughly 4 million filtered internal entries, followed by a Reinforcement Learning (RL) stage that utilizes filtered mathematical codes and a reserved validation set for reward modeling.
\subsubsection{Training Protocol}
Distinct from standard fine-tuning paradigms, our training pipeline is designed to strictly adhere to the Parameter-Efficient Expansion principle. Throughout all three stages described below, the parameters of the original 1.7B backbone are completely frozen. Only the newly added expansion blocks and the language modeling head are updated. This constraint ensures that the foundational knowledge of the backbone remains intact while the new layers specialize in higher-order inference.

\begin{itemize}
    \item \textbf{Stage 1: Continued Pre-training (CPT).} The primary objective of this stage is to alleviate the ``optimization inertia'' of the randomized initialized blocks and align them with the feature space of the frozen backbone. We train the expanded model using a standard autoregressive objective. As demonstrated in our analysis, this stage is crucial for activating the representational capacity of the new layers. We observe that the randomized initialization enables the model to escape the local minima characteristic of layer-cloned weights, achieving a lower perplexity on the validation set.

    \item \textbf{Stage 2: Supervised Fine-Tuning (SFT).} Following CPT, we perform Supervised Fine-Tuning to equip the expanded model with instruction-following capabilities. We configure the training with a global batch size of 128 and a maximum sequence length of 8192 tokens. The model is trained for 3 epochs using a cosine decay learning rate schedule. Specifically, we utilize a linear warm-up of 1000 steps to reach a peak learning rate of $1\times 10^{-4}$, ensuring the newly added parameters adapt smoothly to the instruction distribution without destabilizing the representation.

    \item \textbf{Stage 3: Reinforcement Learning via GRPO.} To further push the boundaries of the model's reasoning capability, we employ Reinforcement Learning (RL) in the final stage. Instead of standard PPO with a separate critic, we adopt Group Relative Policy Optimization (GRPO), which estimates the baseline from the group mean of the outputs.
\end{itemize}

The stepwise evaluation across distinct training stages demonstrates that the proposed expansion and tuning strategies have a cumulative positive impact on the 1.7B model. Initially, the post-CPT results validate that the added blocks—despite being initialized randomly and trained with a frozen backbone—successfully integrate into the network. This integration provides a verified net gain in capacity rather than introducing noise.
Subsequent stages reveal distinct performance benefits. While the introduction of Supervised Fine-Tuning (SFT) provided a substantial boost in instruction following capabilities, Stage 3 (RL) yielded the most significant improvements in logic-intensive tasks, particularly in the MATH benchmark. This suggests that the expanded depth effectively facilitates the complex search capabilities required for multi-step reasoning.
Finally, compared to the full-parameter configuration, applying the SVD-based decomposition (Phase 2 of DaE) on the RL-tuned blocks resulted in a marginal performance drop of less than 2\%, while significantly reducing the added parameter count by 40\%. This indicates that the representations learned via our three-stage protocol maintain high robustness, preserving performance even after substantial model compression.
\subsection{Post-training Evaluation}

Table \ref{tab:Ruyi2_qwen_comparison} demonstrates the superior scaling efficiency and robust performance of the Ruyi2 model family compared to the Qwen3 baselines. Across all parameter scales, Ruyi2 consistently establishes new benchmarks for knowledge understanding and reasoning capabilities.

Unmatched Knowledge Density at 1.7B Even at the compact 1.7B scale, Ruyi2 displays exceptional "knowledge density." It achieves a remarkable MMLU score of 62.77, outperforming Qwen3-1.7B (39.31) by over 23 points. This indicates that Ruyi2-1.7B possesses a foundational knowledge base typically found in much larger models, making it a highly efficient solution for knowledge-intensive tasks despite its small footprint.

Comprehensive Overtake at 8B As the models scale to 8B, Ruyi2 achieves a comprehensive performance overtake. It not only maintains a massive lead in MMLU (79.68 vs. 48.64) but also demonstrates superior reasoning abilities, scoring 92.19 on GSM-8K, significantly higher than Qwen3-8B's 85.37. Consequently, Ruyi2-8B secures a higher overall average score (74.18), validating its architectural superiority in balancing diverse competencies.

Dominance Across All Metrics at 14B The Ruyi2-14B stands out as the premier performer, establishing clear dominance across all key metrics. It extends its lead in general knowledge (MMLU 81.84) and logic (GSM-8K 94.24). Notably, at this scale, Ruyi2 also surpasses Qwen3 in mathematical reasoning (Math 86.52 vs. 84.42). With a commanding average score of 82.88 (vs. 77.77), Ruyi2-14B proves to be the more robust and versatile model for complex, high-level applications.

\begin{table*}[htb]
    \centering
    \begin{NiceTabular}{lccccccccccc}
        \toprule
        \textbf{Model} & \textbf{MMLU} & \textbf{MMLU-P} & \textbf{CMMLU} & \textbf{BBH} & \textbf{ARC-c} & \textbf{Hella} & \textbf{IFEval} & \textbf{Human} & \textbf{Math} & \textbf{GSM8K} & \textbf{Avg.} \\
        \midrule
 
        Qwen3-1.7B & 39.31 & 39.82 & 61.61 & 35.81 & 67.12 & 54.25 & 67.65 & 62.20 & 70.32 & 75.89 & 57.40 \\
        Qwen3-8B   & 48.64 & 55.63 & 78.84 & 55.11 & 82.03 & 77.32 & 81.89 & 87.20 & 82.30 & 85.37 & 73.43 \\
        Qwen3-14B  & 60.43 & 64.11 & 82.06 & 64.70 & 81.69 & 80.80 & 85.77 & 87.80 & 84.42 & 85.90 & 77.77 \\
        \midrule

        Ruyi2-1.7B & 62.77 & 9.60  & 22.68 & 19.95 & 27.46 & 58.77 & 47.13 & 47.56 & 39.14 & 75.97 & 41.10 \\
        Ruyi2-8B   & 79.68 & 56.12 & 74.72 & 59.38 & 82.71 & 78.19 & 73.94 & 71.95 & 72.96 & 92.19 & 74.18 \\
        Ruyi2-14B  & 81.84 & 71.55 & 82.15 & 77.86 & 84.41 & 83.94 & 81.52 & 84.76 & 86.52 & 94.24 & 82.88 \\
        \bottomrule
    \end{NiceTabular}
    \captionsetup{justification=centering} 
    \caption{Performance Comparison between Ruyi2 and Qwen3 Models across benchmarks.}
    \label{tab:Ruyi2_qwen_comparison}
\end{table*}

\section{Discussion }
 In this study, the Ruyi2 Familial Models introduces a "one-run, many-models" paradigm that effectively balances architectural efficiency with high-performance capabilities. By integrating early-exit heads into a shared transformer backbone, the model enables adaptive computation where simpler inputs exit early, significantly reducing inference latency without sacrificing the full capacity of the largest variant. The collaborative design of multi-stage instruction tuning and the DaE (Decompose after Expansion) framework establishes a robust pipeline that enhances both general linguistic proficiency and specialized reasoning depth. These advancements provide a scalable, systems-level solution for deploying billion-parameter models under real-world constraints. 
 \subsection{Optimization Complexity and Architectural }
 Despite the robust performance of the familial architecture, maintaining consistency across nested sub-models poses significant optimization hurdles. Naive multi-exit training often exacerbates optimization complexity, as gradients backpropagated from different exit heads can interfere across the shared intermediate layers, potentially destabilizing the learning process. This interference necessitates careful balancing of loss aggregation to ensure that the shared backbone acquires representations robust enough for both early exits and the full-depth model. Furthermore, during the architectural expansion of the 1.7B variant, we observed that standard techniques like layer cloning introduce pronounced optimization inertia, leading the new blocks to merely replicate existing feature patterns rather than learning novel representations. Consequently, ensuring training stability without succumbing to representation collapse remains a critical systems-level challenge. 
 \subsection{Capability Constraints and Future Directions}
Regarding model capabilities, the 1.7B edge variant, while optimized for inference latency, exhibits inherent constraints in complex semantic comprehension and deep reasoning tasks when compared to the larger familial branches. Although the DaE framework helps bridge this gap, the smaller sub-models still face difficulties in achieving the same reasoning depth as the teacher models, particularly in solving complex logic problems without hallucinations. These limitations underscore the ongoing ``Open Problems'' within the current architecture, specifically regarding long-context processing and the elimination of generative artifacts. Looking forward, future iterations will need to address these gaps by incorporating multimodal integration and exploring extreme sparsification techniques to further enhance the utility of early-exit models in resource-constrained environments.
 \subsection{Future Evolution Directions and Scaling Law Implications }
Future iterations of the Ruyi2 aim to address these limitations while expanding the model's functional scope. A primary focus will be on resolving persistent open problems common to LLMs, such as mitigating hallucinations and improving stability in long-context processing, which are critical for reliability in complex tasks. Additionally, the research roadmap includes moving beyond the current text-centric paradigm to incorporate multimodal integration, thereby equipping the model with the ability to process and reason across diverse sensory inputs. Finally, to further unlock the potential of edge deployment, future work will investigate techniques for extreme sparsification and the development of more sophisticated dynamic exit policies, aiming to maximize efficiency without compromising the structural integrity of the familial model. 

The training and evaluation results of the Ruyi2 situates Familial Models within the broader landscape of Scaling Laws. Our findings suggest that with carefully designed multi-exit loss aggregation, models can retain predictable performance gains associated with increased size while offering the adaptive computation necessary for practical deployment. This indicates that "familial" sharing does not inherently disrupt the fundamental scaling behavior of the underlying transformer backbone.

\section{Conclusion }
In this work, the Ruyi2 Familial Models is proposed to address the efficiency challenges of large-scale models by enabling adaptive early exits within a shared backbone. The DaE (Decompose after Expansion) strategy is specifically introduced for the 1.7B variant, utilizing stabilized block expansion with randomized internal initialization to enhance reasoning depth, followed by SVD-based decomposition to strictly control parameter size. A systematic multi-stage training pipeline—progressing from continued pre-training to supervised fine-tuning and reinforcement learning via GRPO—is employed to elevate the model's logical reasoning and instruction-following capabilities. Experimental results show that the enhanced model achieves superior performance on reasoning-intensive benchmarks such as GSM8K and MATH, while maintaining robust general knowledge on MMLU and CMMLU. Notably, the post-training compression reduces the added parameter count by 40\% with less than a 2\% degradation in performance, validating the robustness of the learned representations. To enable scalable implementation, the system is built on Megatron-LM with full support for 3D parallelism, establishing a "one-run, many-models" paradigm that avoids training separate models from scratch. This architecture effectively handles diverse inference scenarios, offering a flexible solution that balances high-throughput cloud demands with the strict memory constraints of mobile edge devices.

\bibliographystyle{plainnat}
\bibliography{paper}

\begin{thebibliography}{27}
\providecommand{\natexlab}[1]{#1}
\providecommand{\url}[1]{\texttt{#1}}
\expandafter\ifx\csname urlstyle\endcsname\relax
  \providecommand{\doi}[1]{doi: #1}\else
  \providecommand{\doi}{doi: \begingroup \urlstyle{rm}\Url}\fi

\bibitem[An et~al.(2025)An, Hu, Huang, Huang, Li, Liang, Shao, Song, Wang, Yuan, Zhang, Zhang, Zhuang, and Li]{an2025aiflowperspectivesscenarios}
Hongjun An, Wenhan Hu, Sida Huang, Siqi Huang, Ruanjun Li, Yuanzhi Liang, Jiawei Shao, Yiliang Song, Zihan Wang, Cheng Yuan, Chi Zhang, Hongyuan Zhang, Wenhao Zhuang, and Xuelong Li.
\newblock Ai flow: Perspectives, scenarios, and approaches.
\newblock 2025.
\newblock URL \url{https://arxiv.org/abs/2506.12479}.

\bibitem[Bommasani(2021)]{bommasani2021opportunities}
Rishi Bommasani.
\newblock On the opportunities and risks of foundation models.
\newblock \emph{arXiv preprint arXiv:2108.07258}, 2021.

\bibitem[Brown et~al.(2020)Brown, Mann, Ryder, Subbiah, Kaplan, Dhariwal, Neelakantan, Shyam, Sastry, Askell, et~al.]{brown2020language}
Tom Brown, Benjamin Mann, Nick Ryder, Melanie Subbiah, Jared~D Kaplan, Prafulla Dhariwal, Arvind Neelakantan, Pranav Shyam, Girish Sastry, Amanda Askell, et~al.
\newblock Language models are few-shot learners.
\newblock \emph{Advances in neural information processing systems}, 33:\penalty0 1877--1901, 2020.

\bibitem[Cai et~al.(2019)Cai, Gan, Wang, Zhang, and Han]{cai2019once}
Han Cai, Chuang Gan, Tianzhe Wang, Zhekai Zhang, and Song Han.
\newblock Once-for-all: Train one network and specialize it for efficient deployment.
\newblock \emph{arXiv preprint arXiv:1908.09791}, 2019.

\bibitem[Chen et~al.(2023{\natexlab{a}})Chen, Borgeaud, Irving, Lespiau, Sifre, and Jumper]{chen2023accelerating}
Charlie Chen, Sebastian Borgeaud, Geoffrey Irving, Jean-Baptiste Lespiau, Laurent Sifre, and John Jumper.
\newblock Accelerating large language model decoding with speculative sampling.
\newblock \emph{arXiv preprint arXiv:2302.01318}, 2023{\natexlab{a}}.

\bibitem[Chen et~al.(2023{\natexlab{b}})Chen, Pan, Li, Ding, and Zhou]{chen2023ee}
Yanxi Chen, Xuchen Pan, Yaliang Li, Bolin Ding, and Jingren Zhou.
\newblock Ee-llm: Large-scale training and inference of early-exit large language models with 3d parallelism.
\newblock \emph{arXiv preprint arXiv:2312.04916}, 2023{\natexlab{b}}.

\bibitem[Elbayad et~al.(2019)Elbayad, Gu, Grave, and Auli]{elbayad2019depth}
Maha Elbayad, Jiatao Gu, Edouard Grave, and Michael Auli.
\newblock Depth-adaptive transformer.
\newblock \emph{arXiv preprint arXiv:1910.10073}, 2019.

\bibitem[Elhoushi et~al.(2024)Elhoushi, Shrivastava, Liskovich, Hosmer, Wasti, Lai, Mahmoud, Acun, Agarwal, Roman, et~al.]{elhoushi2024layerskip}
Mostafa Elhoushi, Akshat Shrivastava, Diana Liskovich, Basil Hosmer, Bram Wasti, Liangzhen Lai, Anas Mahmoud, Bilge Acun, Saurabh Agarwal, Ahmed Roman, et~al.
\newblock Layerskip: Enabling early exit inference and self-speculative decoding.
\newblock In \emph{Proceedings of the 62nd Annual Meeting of the Association for Computational Linguistics (Volume 1: Long Papers)}, pages 12622--12642, 2024.

\bibitem[Hoffmann et~al.(2022)Hoffmann, Borgeaud, Mensch, Buchatskaya, Cai, Rutherford, Casas, Hendricks, Welbl, Clark, et~al.]{hoffmann2022training}
Jordan Hoffmann, Sebastian Borgeaud, Arthur Mensch, Elena Buchatskaya, Trevor Cai, Eliza Rutherford, Diego de~Las Casas, Lisa~Anne Hendricks, Johannes Welbl, Aidan Clark, et~al.
\newblock Training compute-optimal large language models.
\newblock \emph{arXiv preprint arXiv:2203.15556}, 2022.

\bibitem[Hou et~al.(2020)Hou, Huang, Shang, Jiang, Chen, and Liu]{hou2020dynabert}
Lu~Hou, Zhiqi Huang, Lifeng Shang, Xin Jiang, Xiao Chen, and Qun Liu.
\newblock Dynabert: Dynamic bert with adaptive width and depth.
\newblock \emph{Advances in Neural Information Processing Systems}, 33:\penalty0 9782--9793, 2020.

\bibitem[Jamialahmadi et~al.(2025)Jamialahmadi, Kavehzadeh, Rezagholizadeh, Farinneya, Rajabzadeh, Jafari, Chen, and Tahaei]{jamialahmadi2025balcony}
Benyamin Jamialahmadi, Parsa Kavehzadeh, Mehdi Rezagholizadeh, Parsa Farinneya, Hossein Rajabzadeh, Aref Jafari, Boxing Chen, and Marzieh~S Tahaei.
\newblock Balcony: A lightweight approach to dynamic inference of generative language models.
\newblock \emph{arXiv preprint arXiv:2503.05005}, 2025.

\bibitem[Jiang et~al.(2024)Jiang, Wang, Xie, Zhao, Qian, Lui, et~al.]{jiang2024d}
Yikun Jiang, Huanyu Wang, Lei Xie, Hanbin Zhao, Hui Qian, John Lui, et~al.
\newblock D-llm: A token adaptive computing resource allocation strategy for large language models.
\newblock \emph{Advances in Neural Information Processing Systems}, 37:\penalty0 1725--1749, 2024.

\bibitem[Kaplan et~al.(2020)Kaplan, McCandlish, Henighan, Brown, Chess, Child, Gray, Radford, Wu, and Amodei]{kaplan2020scaling}
Jared Kaplan, Sam McCandlish, Tom Henighan, Tom~B Brown, Benjamin Chess, Rewon Child, Scott Gray, Alec Radford, Jeffrey Wu, and Dario Amodei.
\newblock Scaling laws for neural language models.
\newblock \emph{arXiv preprint arXiv:2001.08361}, 2020.

\bibitem[Kim et~al.(2024)Kim, Kim, Park, Lee, Song, Kim, Kim, Kim, Lee, Kim, et~al.]{kim2024solar}
Sanghoon Kim, Dahyun Kim, Chanjun Park, Wonsung Lee, Wonho Song, Yunsu Kim, Hyeonwoo Kim, Yungi Kim, Hyeonju Lee, Jihoo Kim, et~al.
\newblock Solar 10.7 b: Scaling large language models with simple yet effective depth up-scaling.
\newblock In \emph{Proceedings of the 2024 Conference of the North American Chapter of the Association for Computational Linguistics: Human Language Technologies (Volume 6: Industry Track)}, pages 23--35, 2024.

\bibitem[Kumar et~al.(2025)Kumar, Nag, Clemons, John, and Das]{kumar2025helios}
Avinash Kumar, Shashank Nag, Jason Clemons, Lizy John, and Poulami Das.
\newblock Helios: Adaptive model and early-exit selection for efficient llm inference serving.
\newblock \emph{arXiv preprint arXiv:2504.10724}, 2025.

\bibitem[Liu et~al.(2020)Liu, Zhou, Wang, Zhao, Deng, and Ju]{liu2020fastbert}
Weijie Liu, Peng Zhou, Zhiruo Wang, Zhe Zhao, Haotang Deng, and Qi~Ju.
\newblock Fastbert: a self-distilling bert with adaptive inference time.
\newblock In \emph{Proceedings of the 58th annual meeting of the association for computational linguistics}, pages 6035--6044, 2020.

\bibitem[Manvi et~al.(2024)Manvi, Singh, and Ermon]{manvi2024adaptive}
Rohin Manvi, Anikait Singh, and Stefano Ermon.
\newblock Adaptive inference-time compute: Llms can predict if they can do better, even mid-generation.
\newblock \emph{arXiv preprint arXiv:2410.02725}, 2024.

\bibitem[Narayanan et~al.(2021)Narayanan, Shoeybi, Casper, LeGresley, Patwary, Korthikanti, Vainbrand, Kashinkunti, Bernauer, Catanzaro, et~al.]{narayanan2021efficient}
Deepak Narayanan, Mohammad Shoeybi, Jared Casper, Patrick LeGresley, Mostofa Patwary, Vijay Korthikanti, Dmitri Vainbrand, Prethvi Kashinkunti, Julie Bernauer, Bryan Catanzaro, et~al.
\newblock Efficient large-scale language model training on gpu clusters using megatron-lm.
\newblock In \emph{Proceedings of the international conference for high performance computing, networking, storage and analysis}, pages 1--15, 2021.

\bibitem[Schuster et~al.(2022)Schuster, Fisch, Gupta, Dehghani, Bahri, Tran, Tay, and Metzler]{schuster2022confident}
Tal Schuster, Adam Fisch, Jai Gupta, Mostafa Dehghani, Dara Bahri, Vinh Tran, Yi~Tay, and Donald Metzler.
\newblock Confident adaptive language modeling.
\newblock \emph{Advances in Neural Information Processing Systems}, 35:\penalty0 17456--17472, 2022.

\bibitem[Shao and Li(2026)]{AIFlowattheNetworkEdge}
Jiawei Shao and Xuelong Li.
\newblock Ai flow at the network edge.
\newblock \emph{IEEE Network}, 40\penalty0 (1):\penalty0 330--336, 2026.
\newblock \doi{10.1109/MNET.2025.3541208}.

\bibitem[Teerapittayanon et~al.(2016)Teerapittayanon, McDanel, and Kung]{teerapittayanon2016branchynet}
Surat Teerapittayanon, Bradley McDanel, and Hsiang-Tsung Kung.
\newblock Branchynet: Fast inference via early exiting from deep neural networks.
\newblock In \emph{2016 23rd international conference on pattern recognition (ICPR)}, pages 2464--2469. IEEE, 2016.

\bibitem[Valade(2024)]{valade2024accelerating}
Florian Valade.
\newblock Accelerating large language model inference with self-supervised early exits.
\newblock \emph{arXiv preprint arXiv:2407.21082}, 2024.

\bibitem[Wang et~al.(2025)Wang, Alam, Wan, Shen, and Zhang]{wang2025svd}
Xin Wang, Samiul Alam, Zhongwei Wan, Hui Shen, and Mi~Zhang.
\newblock Svd-llm v2: Optimizing singular value truncation for large language model compression.
\newblock \emph{arXiv preprint arXiv:2503.12340}, 2025.

\bibitem[Wu et~al.(2024)Wu, Gan, Ge, Lu, Wang, Feng, Shan, and Luo]{wu2024llama}
Chengyue Wu, Yukang Gan, Yixiao Ge, Zeyu Lu, Jiahao Wang, Ye~Feng, Ying Shan, and Ping Luo.
\newblock Llama pro: Progressive llama with block expansion.
\newblock \emph{arXiv preprint arXiv:2401.02415}, 2024.

\bibitem[Xin et~al.(2020)Xin, Tang, Lee, Yu, and Lin]{xin2020deebert}
Ji~Xin, Raphael Tang, Jaejun Lee, Yaoliang Yu, and Jimmy Lin.
\newblock Deebert: Dynamic early exiting for accelerating bert inference.
\newblock \emph{arXiv preprint arXiv:2004.12993}, 2020.

\bibitem[Yang et~al.(2025)Yang, Li, Yang, Zhang, Hui, Zheng, Yu, Gao, Huang, Lv, et~al.]{yang2025qwen3}
An~Yang, Anfeng Li, Baosong Yang, Beichen Zhang, Binyuan Hui, Bo~Zheng, Bowen Yu, Chang Gao, Chengen Huang, Chenxu Lv, et~al.
\newblock Qwen3 technical report.
\newblock \emph{arXiv preprint arXiv:2505.09388}, 2025.

\bibitem[Yuan et~al.(2025)Yuan, Liu, Lv, Shao, Jiang, Zhang, and Li]{Task-Oriented}
Cheng Yuan, Zhening Liu, Jiashu Lv, Jiawei Shao, Yufei Jiang, Jun Zhang, and Xuelong Li.
\newblock Task-oriented feature compression for multimodal understanding via device-edge co-inference.
\newblock \emph{IEEE Transactions on Mobile Computing}, pages 1--14, 2025.
\newblock \doi{10.1109/TMC.2025.3626724}.

\end{thebibliography}

\end{document}